\documentclass[10pt,twocolumn,letterpaper]{article}

\usepackage[pagenumbers]{wacv} 

\definecolor{wacvblue}{rgb}{0.21,0.49,0.74}
\usepackage[pagebackref,breaklinks,colorlinks,allcolors=wacvblue]{hyperref}
\usepackage{multirow,multicol,placeins} 
\def\wacvPaperID{1033} 
\def\confName{WACV}
\def\confYear{2027}

\title{High-Fidelity Video Quality Assessment with VQA-Specific Saliency}

\author{
Hakan Emre Gedik$^{1}$, Shashank Gupta$^{1}$, Alan Bovik$^{1,2}$ \\
$^{1}$ The University of Texas at Austin, $^{2}$ University of Colorado Boulder \\
{\tt\small
\{hakan.gedik, shashank.gupta\}@utexas.edu \;
Alan.Bovik@colorado.edu
}
}
\begin{document}
\maketitle
\begin{abstract}
No-reference video quality assessment (NR VQA) has recently seen promising progress with deep learning. However, video data is inherently large, and processing them with deep models incurs high computational cost. This challenge is particularly acute in VQA, where preserving original-resolution cues and dense temporal information is critical for accuracy. Existing efficiency-driven preprocessing strategies, such as fragmenting, reduce computation but alter the input data distribution, limiting effective reuse of pretrained video foundation models (ViFMs). To address these challenges, we propose \textbf{H}igh-\textbf{F}idelity \textbf{V}ideo \textbf{Q}uality \textbf{A}ssessment (\textbf{HFVQA}), a framework built on fixed-size spatio-temporal (ST) patches that is fully compatible with pretrained ViFMs. HFVQA samples ST patches across multiple scales, including the original resolution, with minimal temporal subsampling to preserve low-level quality cues and semantic context. To limit computation, HFVQA introduces a lightweight auxiliary network trained end-to-end with the ViFM encoder to learn \textit{VQA-specific saliency}. Distilled directly from quality supervision, this saliency captures task-specific importance patterns, reflecting that video quality perception is dominated by a small subset of spatio-temporal regions. By combining high-fidelity spatio-temporal cues with learned, task-specific saliency, HFVQA achieves SOTA performance on standard NR VQA benchmarks while processing as little as 12\% of candidate ST patches, making high-fidelity ViFM-based VQA computationally tractable.
\end{abstract}

\section{Introduction}
\label{sec:intro}
No-reference video quality assessment (NR-VQA) is a challenging problem due to the diversity of visual content and real-world distortions. Early approaches relied on handcrafted natural scene statistics (NSS) features \cite{brisque, bliinds, viideo, early_st, early_codec, tlvqm, ugcvqa}, but such methods struggle to generalize beyond limited distortion models. More recently, deep learning–based methods \cite{vmeon, pro_nrvqa_ugc, conviq, patchvq, deepnrvqa_icip, fastvqa, fastervqa, mvqa, mbvqa, zoomvqa} have demonstrated strong performance by learning data-driven quality representations. However, video data is inherently large, and naively processing dense spatio-temporal signals with deep models incurs substantial computational cost. Therefore, how video data is processed has become a central design challenge for enabling practical VQA deployment \cite{fastvqa, fastervqa}.

Unlike recognition-centric vision tasks, VQA depends not only on high-level semantics but also on low-level characteristics of visual signals. While aggressive spatial or temporal subsampling is often acceptable for semantic understanding, such strategies risk discarding quality-critical cues. For example, spatial downsampling can obscure artifacts related to sharpness or noise, while temporal subsampling may mask flicker and motion-related distortions. Consequently, input processing is a particularly critical design choice in VQA, as it must balance computational efficiency with the preservation of fine-grained quality cues.

However, some widely used NR VQA benchmarks contain distortions that are largely spatial or slowly varying in time, where heavy temporal subsampling (e.g., 3.2 FPS) incurs little performance degradation \cite{simpleVQA}. Partly for this reason, many SOTA methods \cite{mvqa, kvq, fastervqa, fastvqa, dover} adopt sparse frame sampling to reduce computational cost. Such subsampling, however, limits sensitivity to high-frequency temporal artifacts such as flicker or fast motion, causing these methods to underperform on temporally challenging benchmarks.

Another challenge in deep VQA is the scarcity of annotated data. Subjective quality assessment studies are expensive and time-consuming, severely limiting the size of publicly available datasets. Moreover, supervision is typically restricted to a single scalar mean opinion score (MOS) per video, which provides weak and ambiguous guidance towards learning robust quality representations. These limitations are commonly mitigated through large-scale video pretraining, including supervised training on action recognition and instructional video datasets \cite{kay2017kinetics, soomro2012ucf101, miech2019howto100m}, as well as self-supervised approaches such as masked video modeling \cite{videomae, videomaev2} and video–text contrastive learning \cite{videoprism, videoclip, internvideo2}. Pretraining provides transferable spatio-temporal priors that compensate for limited VQA supervision. However, pretrained video backbones implicitly assume a specific spatio-temporal input structure. Processing strategies that substantially alter this structure can induce distribution shift and undermine the effectiveness of pretraining. For example, fragmentation-based approaches \cite{fastvqa, fastervqa, mvqa} disrupt spatio-temporal continuity learned during pretraining, complicating reuse of pretrained architectures.

To effectively reuse pretrained video foundation models (ViFMs), we process video data by sampling fixed-size \textbf{S}patio-\textbf{T}emporal (\textbf{ST}) patches that preserve the local structure assumed during pretraining. With fixed patch extents, input resolution introduces a trade-off: higher resolutions preserve low-level quality cues but provide limited semantics, while downsampling emphasizes global structure and semantics at the cost of low-level cues. To balance these complementary aspects, ST patches are sampled from both original-resolution videos and their resized variants. To capture fine-grained temporal quality aspects, patches are extracted at a denser temporal sampling than commonly adopted in prior VQA methods. By preserving original-resolution cues and dense temporal information, we achieve \textbf{H}igh-\textbf{F}idelity \textbf{V}ideo \textbf{Q}uality \textbf{A}ssessment (\textbf{HFVQA}), where ST patches serve as the fundamental units processed by a pretrained ViFM encoder.

While preserving low-level quality cues, semantic context, and dense temporal information is desirable for accurate quality assessment, exhaustively encoding all ST patches is computationally prohibitive. However, perceived video quality is typically dominated by a relatively small subset of spatio-temporal regions \cite{discovqa}. We leverage this observation by introducing a lightweight saliency module that identifies quality-relevant regions and that guides the selection of a reduced set of ST patches for encoding, thereby substantially reducing computational cost.

Crucially, the saliency module is trained jointly with the ViFM encoder and produces dense importance maps that are used to select and aggregate ST patch features. Because this saliency is learned end-to-end under VQA supervision and is directly coupled to the encoder representations, it captures task-specific importance patterns tailored to VQA rather than generic visual saliency. We therefore refer to this mechanism as \textbf{VQA-specific saliency}.

Overall, we propose HFVQA, a VQA framework built on fixed-size ST patches that is fully compatible with pretrained ViFMs. HFVQA jointly preserves fine-grained quality cues, semantic context, and dense temporal information through multiscale and dense ST patch sampling. To make HFVQA computationally tractable, a lightweight, jointly trained VQA-specific saliency module selectively encodes only those spatio-temporal regions most critical to perceived video quality, enabling substantial efficiency gains without modifying the underlying video representation. Our main contributions are summarized as follows:

\begin{enumerate}
\item We propose HFVQA, a fragmentation-free VQA framework that processes video as fixed-size ST patches, preserving the local spatio-temporal structure assumed during ViFM pretraining and enabling distribution-faithful reuse of pretrained priors.
\item To make such ViFM-native processing tractable, we introduce a lightweight, end-to-end \emph{VQA-specific saliency} mechanism that identifies quality-critical regions \emph{before} encoding, unlike generic saliency or post-encoding pooling, selecting only a small subset of ST patches.
\item HFVQA attains SOTA on standard NR VQA benchmarks while encoding as little as 12\% of candidate ST patches, with the largest gains on high-resolution and temporally challenging content.
\end{enumerate}
\section{Related Work}
\label{sec:related_work}
Early no-reference VQA methods \cite{brisque, bliinds, viideo, early_st, early_codec, tlvqm, ugcvqa, stgreed} relied on handcrafted features derived from natural scene statistics (NSS), which capture certain perceptual regularities of visual signals. While effective under limited conditions, these methods struggle to generalize across the wide diversity of content and authentic distortions commonly encountered in real-world videos.

Deep learning–based approaches have shown promise in overcoming these limitations. Early methods typically adopted pretrained backbones in an off-the-shelf manner, training lightweight adapters on top of or combining deep features with handcrafted NSS descriptors \cite{vsfa, rapique, gstvqa}. With advances in GPU hardware and the availability of larger datasets, end-to-end training became feasible, enabling backbone adaptation through fine-tuning on VQA datasets and leading to improved performance \cite{vmeon, pro_nrvqa_ugc, conviq, patchvq, deepnrvqa_icip}.

Despite their strong performance, deep learning–based VQA models often incur high computational cost, which is especially problematic in VQA, where preserving original-resolution spatial cues and dense temporal information is critical. To mitigate this issue, the FastVQA family \cite{fastvqa, fastervqa} introduced video fragmenting, a preprocessing strategy that aggressively reduces spatial redundancy while retaining  original-resolution local structures. However, fragmenting has been criticized as obscuring semantic information \cite{dover}, which is also important for quality assessment. Subsequent works such as DOVER \cite{dover}, Zoom-VQA \cite{zoomvqa}, and CLIF-VQA \cite{clifvqa} addressed this limitation by augmenting fragment-based processing with additional branches operating on resized video to recover semantic context. MVQA \cite{mvqa} further refined fragmenting by selectively replacing fragments with patches from resized views. Collectively, these approaches reduce spatial redundancies and enable more efficient temporal modeling.

While these preprocessing strategies substantially reduce computational cost, they do so by significantly altering the input data distribution, which restricts architectural choices and complicates the direct adoption of recent ViFMs \cite{videoprism, internvideo, internvideo2, vjepa2, videomaev2}. This limitation is particularly critical in VQA, where limited supervision is commonly mitigated through large-scale pretraining to transfer general-purpose spatio-temporal priors. Although it is theoretically possible to pretrain ViFMs directly on such preprocessed representations, ViFM pretraining is extremely compute-intensive and often relies on proprietary datasets \cite{videoprism, internvideo2}, making it impractical for most research settings. HFVQA addresses this challenge by preserving the local spatio-temporal structure of the input video through fixed-size ST patches, maintaining compatibility with pretrained ViFMs while mitigating computational cost via learned \emph{VQA-specific saliency}.

Conventional visual saliency has been explored as a guiding signal in quality assessment models \cite{sci, eye_tracking, visual_importance, jnd_saliency, vsi, saliency_review, saliency_patent, saliency_study_vqa}, typically yielding modest but consistent performance gains. However, generic visual saliency does not necessarily align with the importances of spatio-temporal regions for VQA. HFVQA explicitly learns this notion through a lightweight auxiliary network trained end-to-end with MOS supervision, enabling the model to capture task-specific importance patterns. We refer to this learned notion as \emph{VQA-specific saliency}. Recent work such as ReLIQS \cite{reliqs} explores learning quality-aware saliency for image quality assessment, but does not model temporal dependencies in video.

Several existing NR VQA models, such as KVQ \cite{kvq} and related approaches \cite{zoomvqa, saliency_compressed_vqa}, can also be interpreted as implicitly learning region importance from quality supervision. However, these methods typically estimate importance only after feature encoding, for example via importance-weighted pooling. As a result, they cannot prune spatio-temporal regions prior to encoding and therefore offer limited computational savings. By contrast, HFVQA learns VQA-specific saliency before encoding through a lightweight auxiliary network, enabling selective encoding of a small subset of quality-relevant regions and substantially reducing computational cost.
\section{Method}
\label{sec:method}
\begin{figure*}[t]
    \centering
    \includegraphics[width=0.92\linewidth]{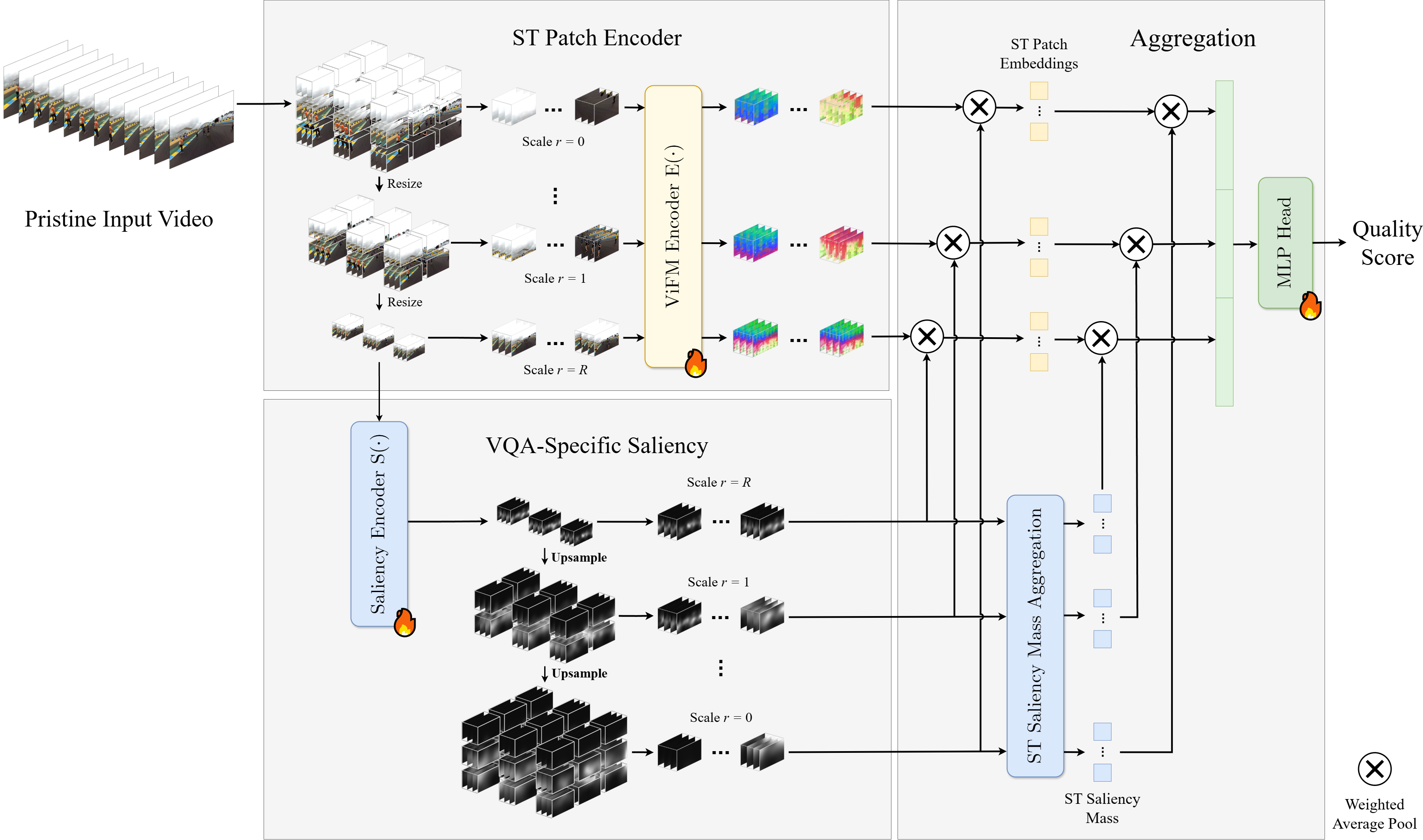}
    \caption{ HFVQA pipeline. Multi-scale ST patches are sampled, with VQA-specific saliency predicted by $S(\cdot)$ at the coarsest scale ($r = R$). Independently encoded patches are aggregated using the resulting saliency volumes.  
    }
    \label{fig:diagram}
\end{figure*}
\subsection{Proposed Framework} 
We propose HFVQA, a high-fidelity VQA framework that preserves original-resolution spatial quality cues and dense temporal information through multiscale, temporally dense, ST patch sampling. HFVQA consists of three core components: (i) a pretrained ViFM that encodes ST patches into quality-aware representations, (ii) a lightweight saliency module that learns VQA-specific saliency over dense spatio-temporal regions, and (iii) an aggregation module that, guided by the learned saliency, fuses multiscale ST patch representations into a final quality prediction. As illustrated in \cref{fig:diagram}, the pipeline proceeds as follows:
\begin{enumerate}
    \item Sample fixed-size ST patches from the video at the original spatial resolution and its resized variants.
    \item Encode each sampled ST patch independently using the pretrained ViFM.
    \item Estimate VQA-specific saliency over spatio-temporal regions via the saliency module.
    \item Aggregate ST patch representations into a quality score under the guidance of the learned saliency.
\end{enumerate}

To be concrete, given an input video $\textbf{x}^{(0)}\in\mathbb{R}^{\text{T}_0 \times 3 \times\text{H}_0 \times \text{W}_0}$, we first construct a set of its spatially resized variants $\{\textbf{x}^{(r)} \in \mathbb{R}^{\text{T}_0 \times 3 \times\text{H}_r \times \text{W}_r}  \}_{r=0}^{\text{R}}$. From each scale, fixed-size ST patches are then sampled according to a predefined strategy, yielding the collection $\{\{\textbf{x}_i^{(r)} \in \mathbb{R}^{T \times 3 \times P \times P} \}_{i=0}^{c_r}  \}_{r=0}^{\text{R}}$, where $T$ denotes the number of frames, and $P$ is the spatial patch size, and $c_r$ is the number of ST patches at scale $r$ . Each sampled ST patch is independently encoded by a pretrained ViFM $E(\cdot)$ to obtain dense spatio-temporal feature representations:
\begin{equation}
    \textbf{e}_i^{(r)} = E(\textbf{x}_i^{(r)}), 
    \qquad
    \textbf{e}_i^{(r)} \in \mathbb{R}^{\frac{T}{T_s} \times \frac{P}{P_s} \times \frac{P}{P_s} \times D}, 
\end{equation}
where $T_s$ and $P_s$ denote the internal temporal and spatial patch sizes of the ViFM encoder, respectively, and $D$ denotes the feature dimensionality. 

In parallel, the saliency module $S(\cdot)$ estimates VQA-specific saliency using the coarsest scale in its original temporal dimension: 
\begin{equation}
    \textbf{s}^{(R)} = S(\textbf{x}^{(R)}), 
    \qquad 
    \textbf{s}^{(R)} \in \mathbb{R}^{\frac{T_R}{T_s} \times \frac{H_R}{P_s} \times \frac{W_R}{P_s}},   
\end{equation}  
where $\textbf{s}^{(R)}$ is normalized to sum to $1$. The saliency module $S(\cdot)$ is implemented as a lightweight spatio-temporal encoder with significantly lower computational cost than $E(\cdot)$. To ensure compatibility with the ViFM feature grid, $S(\cdot)$ is equipped with lightweight adapters that align its temporal and spatial patching with the ViFM patch sizes $T_s$ and $P_s$. To keep the computation resolution-agnostic and tractable, $S(\cdot)$ operates exclusively on a predetermined coarsest scale $(H_R, W_R)$ with a fixed short spatial dimension. Saliency maps at higher spatial scales are obtained by spatially upsampling the coarsest-scale output to ensure spatio-temporal alignment with dense feature tokens across scales: 
\begin{equation}
    \textbf{s}^{(r)} = \text{Upsample}_{\frac{H_r}{P_s}, \frac{W_r}{P_s}}(\textbf{s}^{(R)}),   
    \qquad 
    r=0, \dots, R-1. 
\end{equation}

For the $i$-th ST patch at scale $r$ covering the spatio-temporal grid cells $\mathcal{R}_{i}^{(r)}$, the corresponding saliency sub-volume is extracted from $\mathbf{s}^{(r)}$ as: 
\begin{equation}
    \label{eq:saliency_crop}
    \mathbf{s}_{i}^{(r)} = \mathbf{s}^{(r)}\big|_{\mathcal{R}_{i}^{(r)}},
    \qquad
    \mathbf{s}_{i}^{(r)} \in \mathbb{R}^{\frac{T}{T_s} \times \frac{P}{P_s} \times \frac{P}{P_s}}.
\end{equation}

Using this saliency sub-volume, we first perform within-patch aggregation of dense ViFM features to obtain a single ST patch representation:
\begin{equation}
\label{eq:within_patch_pool}
\begin{gathered}
Z_{i}^{(r)} = \sum_{(t',h',w')\in\mathcal{R}_{i}^{(r)}} \mathbf{s}_{i}^{(r)}(t',h',w'), \\
\mathbf{f}_{i}^{(r)} = \sum_{(t,h,w)\in\mathcal{R}_{i}^{(r)}} \frac{\mathbf{s}_{i}^{(r)}(t,h,w)}{Z_{i}^{(r)}} \;\mathbf{e}_{i}^{(r)}(t,h,w),\quad \mathbf{f}_{i}^{(r)} \in \mathbb{R}^{D}.
\end{gathered}
\end{equation}

Then, ST patch representations are aggregated within each scale using the saliency mass of each patch:
\begin{equation}
\label{eq:patch_saliency_mass}
\begin{split}
\alpha_i^{(r)} &=
\frac{\sum_{(t,h,w)\in\mathcal{R}_{i}^{(r)}} \mathbf{s}_{i}^{(r)}(t,h,w)}
{\sum_{j=1}^{c_r}\sum_{(t,h,w)\in\mathcal{R}_{j}^{(r)}} \mathbf{s}_{j}^{(r)}(t,h,w)}, \\
\mathbf{f}^{(r)} &= \sum_{i=1}^{c_r} \alpha_i^{(r)} \, \mathbf{f}_{i}^{(r)}, \qquad \mathbf{f}^{(r)} \in \mathbb{R}^{D},
\end{split}
\end{equation}
where $\alpha_i^{(r)}$ represents the relative importance of ST patch $i$ among patches at scale $r$ and is normalized to sum to one within the scale. After obtaining the scale-level representations $\mathbf{f}^{(r)}$, we concatenate them and pass the resulting multiscale feature to a regression head with a single hidden layer to predict a scalar quality score:
\begin{equation}
p = \text{Head}\left([\mathbf{f}^{(0)}, \ldots, \mathbf{f}^{(R)}]\right),
\qquad
p \in \mathbb{R},
\end{equation}
where $[\cdot,\cdot]$ denotes concatenation along the feature dimension.

\subsection{Training Objective} 
Following prior work in image and video quality assessment \cite{fastvqa, fastervqa, kvq, mvqa}, we trained HFVQA end-to-end using a combination of margin-ranking and PLCC losses. Given a mini-batch of $N$ videos with ground-truth MOS $\{g_i\}_{i=1}^{N}$ and corresponding predictions $\{p_i\}_{i=1}^{N}$, the margin-ranking loss is defined as
\begin{equation}
    \label{eq:margin-ranking} 
    \mathcal{L}_{\text{MR}} =
        \frac{2}{N(N-1)}\sum_{i<j} \max \Bigl(
            0,\;
            \delta -\,\operatorname{sign}\bigl(g_i - g_j\bigr)
            \bigl(p_i - p_j\bigr)
        \Bigr) 
\end{equation} 
where $\delta$ is the margin hyperparameter. For the same mini-batch, the PLCC loss term is defined as:   
\begin{equation}
\label{eq:plcc}
\mathcal{L}_{\mathrm{PLCC}}
=
1 -
\frac{
    \sum_{i=1}^{N} (g_i - \bar{g})(p_i - \bar{p})
}{
    \sqrt{
        \sum_{i=1}^{N} (g_i - \bar{g})^2
        \sum_{i=1}^{N} (p_i - \bar{p})^2
    }
},
\end{equation}
where $\bar{g}$ and $\bar{p}$ denote the batch means of the ground-truth MOS and predictions, respectively. Our final training objective is: 
\begin{equation}
    \mathcal{L} = \lambda\,\mathcal{L}_{\text{MR}} + (1-\lambda)\,\mathcal{L}_{\text{PLCC}}, 
\end{equation}
where $\lambda \in [0,1]$ controls the trade-off between the two loss terms.
   
\begin{table*}[t]
    \centering
    \footnotesize
    \caption{\textbf{PLCC / SRCC} of compared VQA models. Left: performance under \textbf{LSVQ Pretraining} (cross-dataset entries evaluated over the entire dataset). Right: median performance after task-specific \textbf{Fine-Tuning}. Best entries are \textbf{bold} and second best are \underline{underlined}.}
    \label{tab:main_results}
    \label{tab:finetune_results}
    \setlength{\tabcolsep}{4.5pt}
    \renewcommand{\arraystretch}{0.95}
    \begin{tabular}{l | cccc | cccc}
        \toprule
        \multirow{2}{*}{Methods} & \multicolumn{4}{c|}{\textbf{LSVQ Pretraining}} & \multicolumn{4}{c}{\textbf{Fine-Tuning}} \\
        \cmidrule(lr){2-5} \cmidrule(lr){6-9}
        & \multicolumn{2}{c}{Intra-dataset} & \multicolumn{2}{c|}{Cross-dataset} & \multicolumn{4}{c}{Target Datasets} \\
        \cmidrule(lr){2-3} \cmidrule(lr){4-5} \cmidrule(lr){6-9}
        & \textbf{LSVQ}$_{test}$ & \textbf{LSVQ}$_{1080p}$ & \textbf{KoNViD-1k} & \textbf{LIVE-VQC} & \textbf{KoNViD-1k} & \textbf{LIVE-VQC} & \textbf{YouTube-UGC} & \textbf{LBVD} \\
        \midrule
        TLVQM \cite{tlvqm}      & 0.774 / 0.772 & 0.616 / 0.589 & 0.724 / 0.732 & 0.691 / 0.670 & 0.768 / 0.773 & 0.803 / 0.799 & 0.659 / 0.669 & 0.590 / 0.614 \\
        VIDEVAL \cite{ugcvqa}  & 0.783 / 0.794 & 0.554 / 0.545 & 0.741 / 0.751 & 0.640 / 0.630 & 0.780 / 0.783 & 0.751 / 0.752 & 0.773 / 0.779 &  0.697 / 0.707 \\
        \midrule
        Patch-VQ \cite{patchvq} & 0.828 / 0.827 & 0.739 / 0.711 & 0.795 / 0.791 & 0.807 / 0.770 & 0.786 / 0.791 & 0.837 / 0.827 & -- / --       & -- / -- \\
        BVQA \cite{bvqa}        & 0.854 / 0.852 & 0.782 / 0.771 & 0.837 / 0.834 & 0.824 / 0.816 & 0.836 / 0.834 & 0.842 / 0.831 & 0.819 / 0.831 & \underline{0.887} / \underline{0.891} \\
        VSFA \cite{vsfa}        & 0.796 / 0.801 & 0.704 / 0.675 & 0.794 / 0.784 & 0.772 / 0.734 & 0.775 / 0.773 & 0.795 / 0.773 & 0.743 / 0.724 & 0.642 / 0.622 \\
        Fast-VQA \cite{fastvqa} & 0.874 / 0.872 & 0.809 / 0.770 & 0.862 / 0.864 & 0.841 / 0.824 & 0.889 / 0.890 & 0.852 / 0.845 & 0.853 / 0.857 & 0.809 / 0.804 \\
        FasterVQA \cite{fastervqa}& 0.874 / 0.873 & 0.811 / 0.772 & 0.863 / 0.863 & 0.837 / 0.813 & 0.898 / 0.895 & 0.858 / 0.843 & 0.859 / 0.863 & 0.837 / 0.813\\
        DOVER \cite{dover}      & 0.879 / 0.881 & 0.827 / 0.782 & 0.872 / 0.871 & 0.841 / 0.812 & 0.899 / 0.897 & 0.852 / 0.812 & 0.873 / 0.877 & 0.824 / 0.824 \\
        Q-Align \cite{qalign}   & 0.882 / 0.883 & 0.830 / 0.797 & 0.877 / 0.865 & -- / --       & -- / --       & -- / --       & -- / --       & -- / -- \\
        CLiF-VQA \cite{clifvqa} & 0.887 / 0.886 & 0.832 / 0.790 & 0.874 / 0.877 & 0.855 / 0.834 & 0.903 / 0.903 & 0.878 / 0.866 & 0.890 / 0.888 & -- / -- \\
        MBVQA \cite{mbvqa}      & 0.895 / 0.895 & 0.844 / 0.809 & 0.884 / 0.878 & 0.844 / 0.806 & 0.905 / 0.901 & 0.880 / 0.860 & 0.877 / 0.876 & -- / -- \\
        KVQ \cite{kvq}          & 0.897 / 0.896 & \underline{0.846} / \underline{0.814} & \textbf{0.892} / \underline{0.890} & 0.843 / 0.820 & 0.915 / 0.909 & 0.879 / 0.859 & \underline{0.905} / \underline{0.903} & 0.828 / 0.824 \\
        MVQA \cite{mvqa}        & \underline{0.899} / \underline{0.898} & \underline{0.846} / 0.812 & 0.887 / 0.885 & \textbf{0.873} / \underline{0.852} & \underline{0.925} / \underline{0.925} & \underline{0.895} / \underline{0.878} & 0.903 / 0.901 & -- / -- \\
        \midrule
        \textbf{HFVQA}          & \textbf{0.906} / \textbf{0.903} & \textbf{0.873} / \textbf{0.842} & \underline{0.891} / \textbf{0.893} & \underline{0.862} / \textbf{0.856} & \textbf{0.927} / \textbf{0.926} & \textbf{0.915} / \textbf{0.905} & \textbf{0.921} / \textbf{0.919} & \textbf{0.902} / \textbf{0.904} \\
        \bottomrule
    \end{tabular}
\end{table*} 
\section{Experiments}
\label{sec:experiments}
\subsection{Datasets and Implementation Details}
\paragraph{Datasets.} 
We train and evaluate HFVQA on multiple standard NR VQA benchmarks, including
LSVQ \cite{patchvq}, LIVE-VQC \cite{livevqc}, KoNViD-1k \cite{konvid}, LBVD \cite{lbvd} and YouTube-UGC \cite{youtubeugc}. LSVQ contains 38{,}811 videos, LIVE-VQC 585, KoNViD-1k 1{,}200, LBVD 1{,}013, and the latest version of YouTube-UGC 1{,}067. For LSVQ, we follow the official train/test split provided by the dataset. For the remaining benchmarks, we randomly generate training, validation, and test splits using a 70:10:20 ratio 10 times and report the median performance, following common practice in prior work. All datasets consist of real-world SDR videos with authentic distortions and are predominantly captured at a frame rate of 30~FPS.

\paragraph{Implementation Details.}
For the ViFM encoder $E(\cdot)$, we adopt the VideoPrism~\cite{videoprism} ViViT-B \cite{vivit} factorized encoder variant. To keep the saliency module $S(\cdot)$ light, we employ a VideoSwin-T \cite{videoswin} backbone pretrained on ImageNet-1k \cite{imagenet} and Kinetics-400 \cite{kay2017kinetics}. The ViFM encoder uses internal temporal and spatial patch sizes of $T_s=1$ and $P_s=18$, respectively. To achieve dense spatio-temporal alignment with the ViFM feature grid, $S(\cdot)$ is augmented with a lightweight decoder composed of residual blocks with depthwise 3D convolutions. Exact details of the decoder configuration are provided in the Supplementary Material. 

We trained HFVQA end-to-end using the AdamW optimizer \cite{adamw} with a cosine annealing
learning rate schedule and an initial learning rate of $5\times10^{-6}$ and a weight decay of $10^{-3}$ for a total of 12 epochs. To avoid distorting pretrained representations \cite{lpft}, we froze the pretrained backbones during the first 4 epochs and optimized only the adapter modules, including the prediction head and the saliency decoder. All parameters were then unfrozen and jointly fine-tuned over the remaining 8 epochs. A batch size of 16 was used for LSVQ, while a batch size of 8 was used for the remaining datasets.

We use a spatial ST patch size of $P=252$ and a temporal sampling rate of 10 FPS, where each patch spans $T=16$ consecutive frames. ST patches are sampled from three spatial scales: the original resolution, a resized scale with the short edge set to 252, and an intermediate scale whose short edge is the average of the original short edge and 252. During training, four ST patches were randomly sampled at each scale. At evaluation time, we generated a candidate set of ST patches using uniform grid sampling and then pruned this set using the learned saliency. Further details on the evaluation-time patch sampling strategy are provided in \cref{sec:compute_performance}.

All experiments were conducted using 4 NVIDIA H200 GPUs. We set $\lambda=0.5$ to equally weight the margin-ranking and PLCC loss terms. Following standard practice in VQA, we evaluated performance using Spearman’s rank correlation coefficient (SRCC) and Pearson’s linear correlation coefficient (PLCC).

\subsection{Main Results}  
\begin{figure}[t]
    \centering
    \hspace{0.5em}
    \begin{subfigure}[t]{0.46\linewidth}
        \centering
        \includegraphics[width=\linewidth]{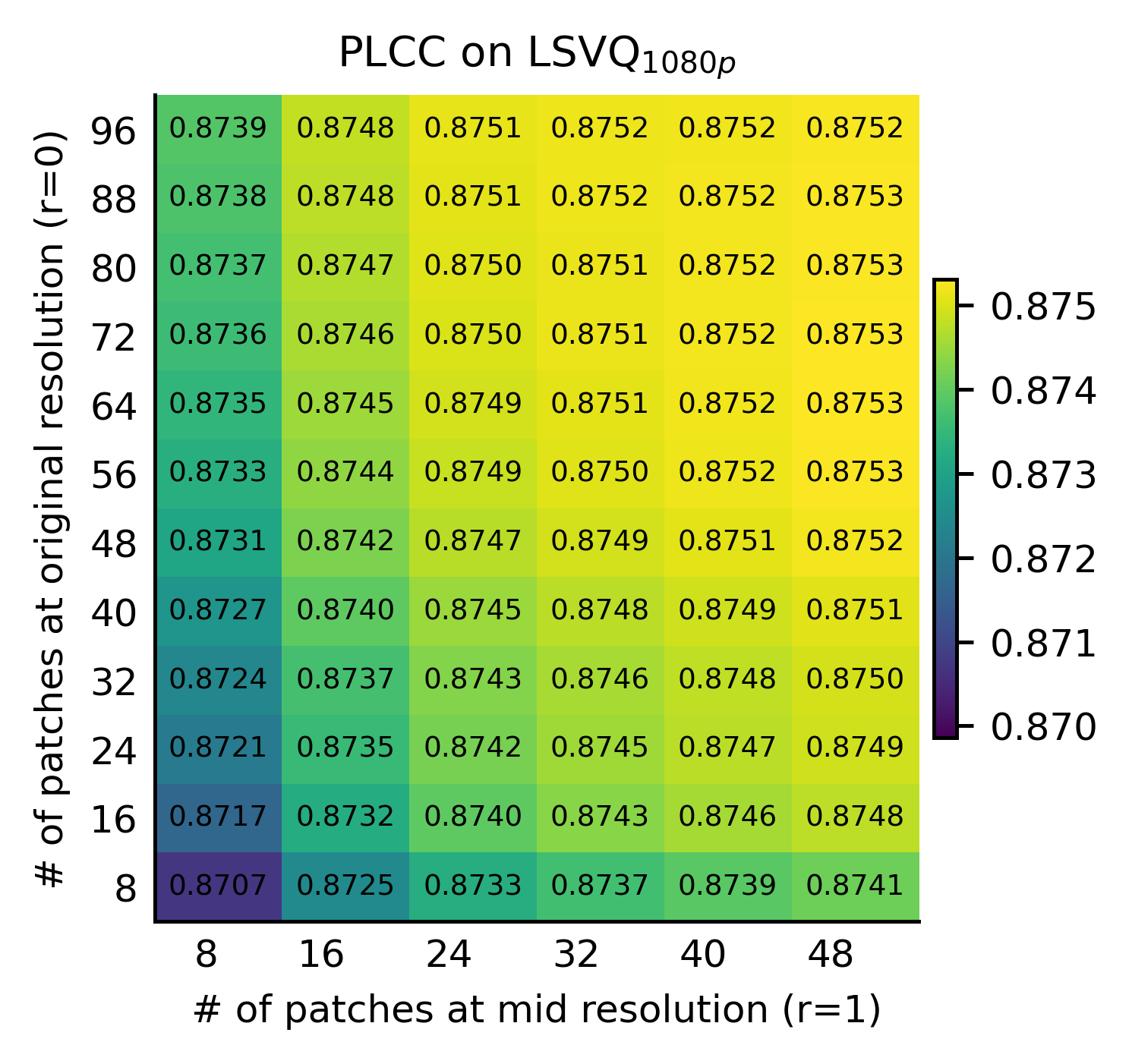}
 
    \end{subfigure}
    \hspace{-0.5em}
    \begin{subfigure}[t]{0.46\linewidth}
        \centering
        \includegraphics[width=\linewidth]{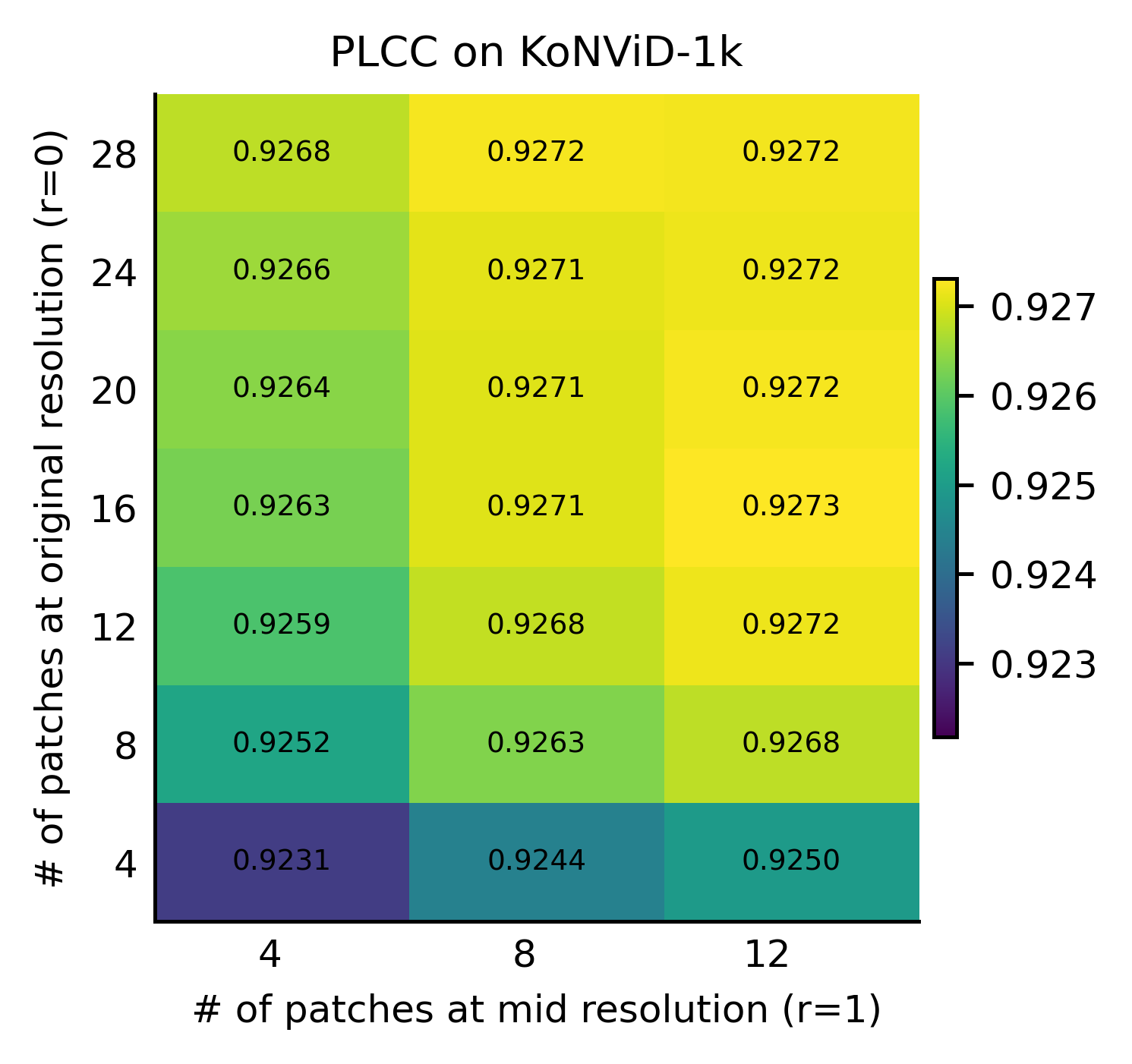}
    \end{subfigure}

    \hspace{0.5em}
    \begin{subfigure}[t]{0.46\linewidth}
        \centering
        \includegraphics[width=\linewidth]{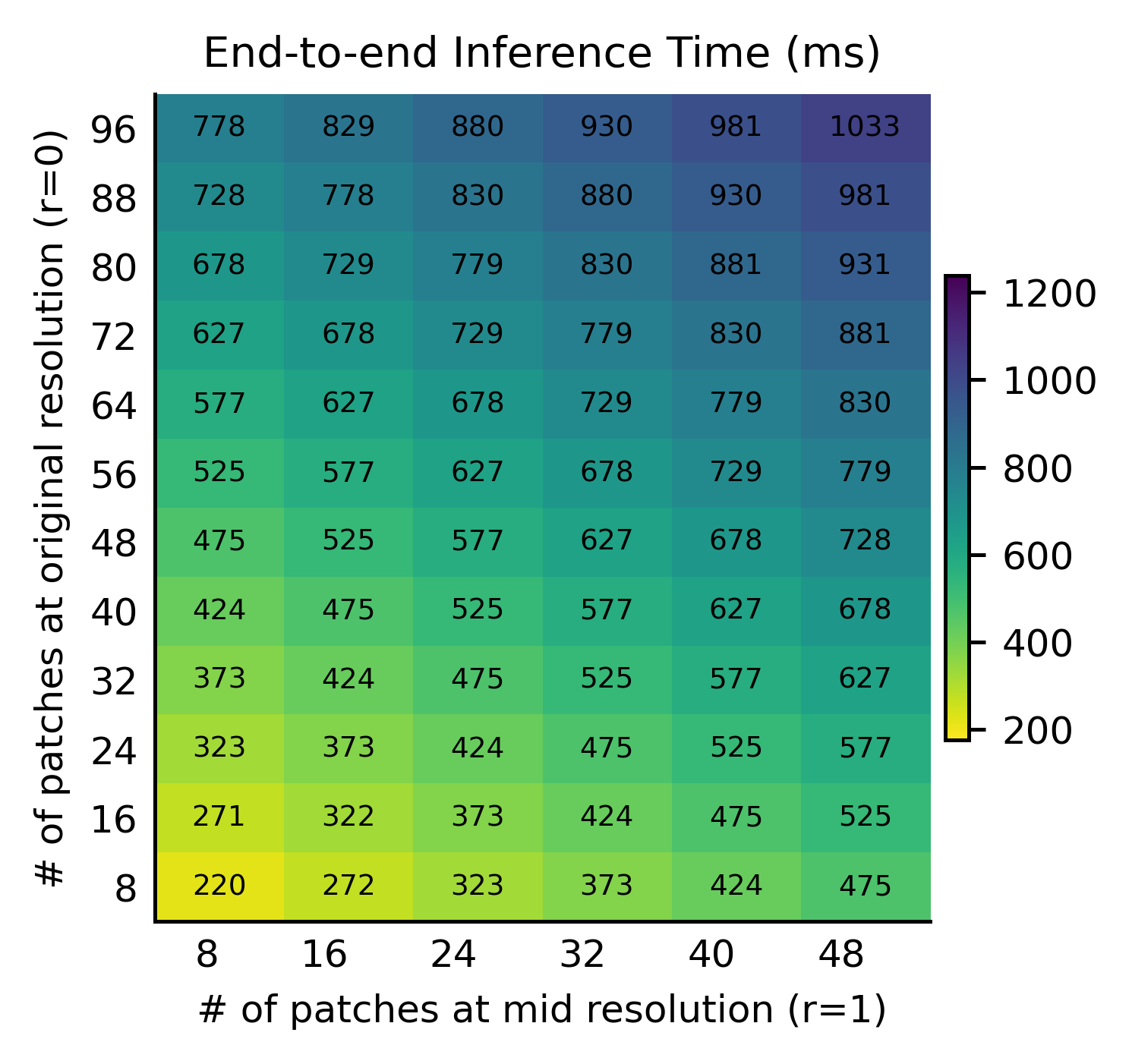}
    \end{subfigure}
    \hspace{0.5em}
    \begin{subfigure}[t]{0.46\linewidth}
        \centering
        \includegraphics[width=\linewidth]{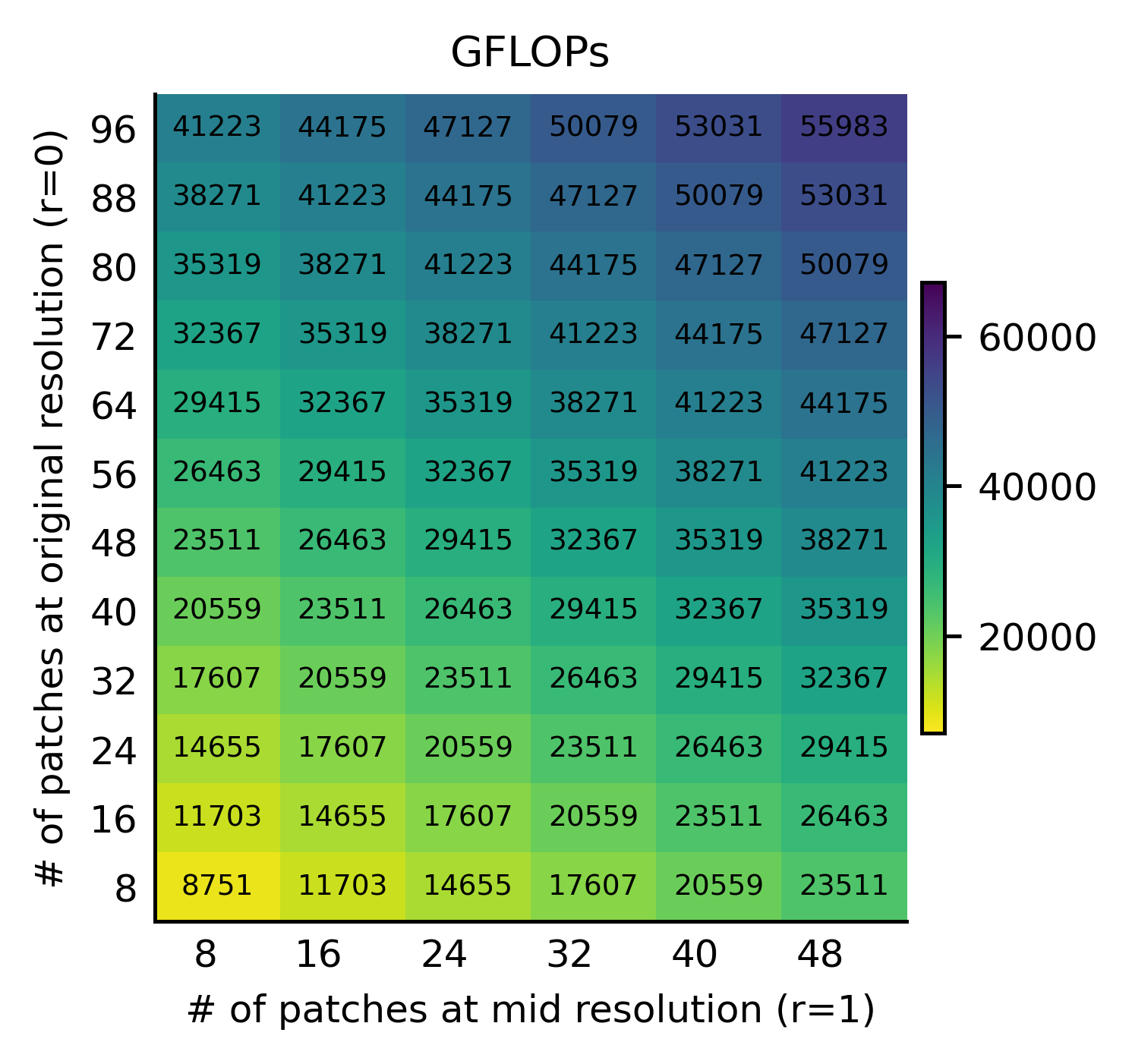}
    \end{subfigure}
    \caption{PLCC on $\text{LSVQ}_{1080p}$ and KoNViD-1k, together with end-to-end inference time and GFLOPs, as a function of the number of ST patches at scales $r=0,1$, with 6 patches fixed at the coarsest scale $r=2$.}
    \label{fig:heatmap}
\end{figure} 
\paragraph{Pretraining on LSVQ.} Most public VQA datasets are relatively small, which limits their suitability for learning robust quality-aware representations from scratch. LSVQ is a notable exception and is therefore commonly used to pretrain VQA models, which are subsequently fine-tuned on smaller benchmarks. In this setting, we train HFVQA on the LSVQ training split and evaluate it on $\text{LSVQ}_{test}$ and $\text{LSVQ}_{1080p}$. To assess cross-dataset generalization, we additionally report performance on the full KoNViD-1k and LIVE-VQC datasets.

As shown in \cref{tab:main_results}, HFVQA substantially outperformed classical NR VQA methods, including TLVQM \cite{tlvqm} and VIDEVAL \cite{ugcvqa}, as well as early deep learning–based baselines such as Patch-VQ~\cite{patchvq}, BVQA \cite{bvqa}, and VSFA \cite{vsfa}. Compared to the fragmentation-based approaches in the FastVQA family \cite{fastvqa,fastervqa}, HFVQA achieved consistent gains, outperforming FasterVQA by 0.032/0.030 PLCC/SRCC on $\text{LSVQ}_{test}$, and by 0.028/0.030 and 0.025/0.043 on KoNViD-1k and LIVE-VQC, respectively, in cross-dataset evaluation.

HFVQA matched or slightly outperformed recent SOTA methods such as KVQ \cite{kvq} and MVQA \cite{mvqa} on benchmarks dominated by low- to mid-resolution content, where aggressive resizing or fragmenting incurs only limited loss of quality-critical information. On the high-resolution $\text{LSVQ}_{1080p}$ benchmark, however, where preserving original-resolution quality cues is essential, HFVQA established a new SOTA with \textbf{0.873}/\textbf{0.842} PLCC/SRCC, beating the previous best MVQA by \textbf{0.027}/\textbf{0.030}.

These results suggest that preserving local spatio-temporal structure while utilizing pretrained ViFMs is effective for VQA, with particularly strong benefits at high resolutions where fine-grained quality cues are most critical.

\paragraph{Fine-tuning.} We further evaluated HFVQA by fine-tuning the LSVQ-pretrained model on smaller NR VQA benchmarks, including LIVE-VQC, KoNViD-1k, LBVD, and YouTube-UGC. Following prior work, HFVQA is initialized from the same LSVQ-pretrained checkpoint used in the previous section. To reduce bias from dataset splits, we performed fine-tuning over 10 random train/validation/test splits and report the median PLCC/SRCC.

As shown in \cref{tab:finetune_results}, HFVQA consistently outperformed prior methods across the evaluated benchmarks. On YouTube-UGC, which features a significant portion of high-resolution videos ($\geq$1080p), HFVQA exceeded KVQ by 0.016/0.016 and MVQA by 0.018/0.018 in PLCC/SRCC, echoing the performance gains observed on high-resolution $\text{LSVQ}_{1080p}$. On the temporally challenging LBVD and LIVE-VQC datasets, HFVQA surpassed KVQ by 0.074/0.080 and FasterVQA by 0.065/0.091 on LBVD, while outperforming MVQA by 0.020/0.027 on LIVE-VQC. 

These results demonstrate that the combination of dense temporal sampling and saliency-guided structural preservation generalizes robustly during fine-tuning, offering clear advantages when modeling high-resolution details and intricate temporal distortions.

\subsection{Compute vs. Performance Tradeoff} 
\label{sec:compute_performance}
\begin{figure*}[t]
    \centering
    \includegraphics[width=0.84\linewidth]{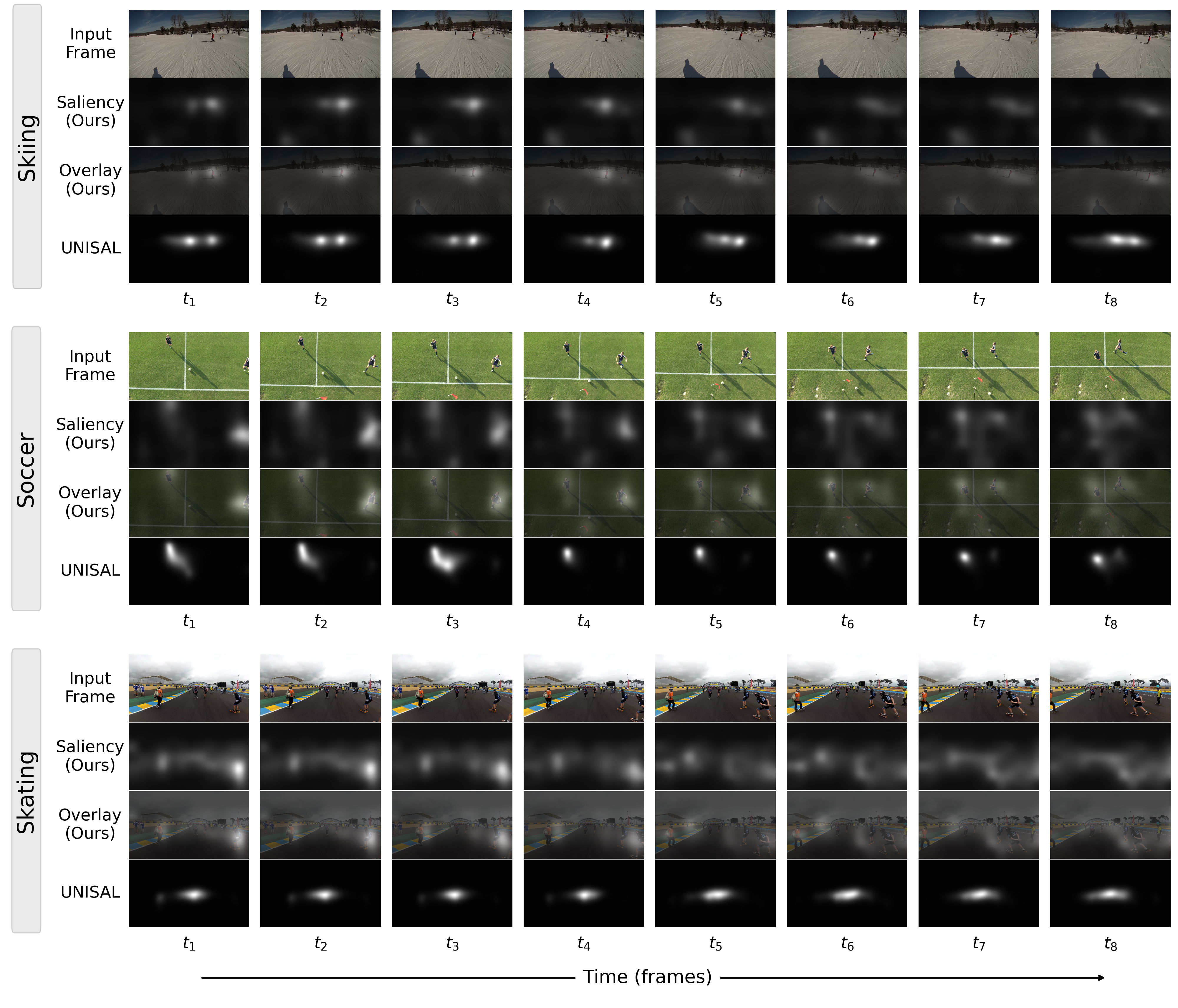}
    \caption{ Consecutive video frames (top), learned VQA-specific saliency maps (second), overlays (third), and comparison with UNISAL (bottom). Best viewed zoomed in.    
    }
    \label{fig:qualitative}
\end{figure*}
We reduced the computational cost of HFVQA by limiting the number of ST patches encoded by the ViFM, guided by the learned VQA-specific saliency. Given an input video, we first constructed a candidate set of non-overlapping ST patches that densely covers the video. Each patch was assigned a saliency score by summing its corresponding saliency volume, and the top-$k_r$ most salient patches were selected at each scale $r$ for encoding.

Since ST patches were processed by a compute-intensive ViFM encoder, the number of encoded patches determined the trade-off between performance and efficiency. To analyze this trade-off and assess the effectiveness of the learned VQA-specific saliency, we visualized performance and efficiency metrics as a function of the number of encoded patches in \cref{fig:heatmap}. In all experiments, the number of patches at the coarsest scale ($r=2$) was fixed to 6, while the numbers of patches at the original resolution ($r=0$) and mid-scale ($r=1$) were varied.

Specifically, we report PLCC on the high-resolution $\text{LSVQ}_{1080p}$ dataset and the lower-resolution KoNViD-1k dataset, together with end-to-end inference time and GFLOPs. As shown in \cref{fig:heatmap}, performances on both datasets saturated rapidly as the number of encoded patches was increased, whereas inference time and GFLOPs grew approximately linearly with the total number of encoded patches.

This behavior indicates that using only a small subset of spatio-temporal regions was sufficient to capture most quality-relevant information, and that the learned VQA-specific saliency was effective at identifying these regions. For example, a $1080\times1920$ video with 96 frames yielded 258 candidate ST patches under dense coverage. HFVQA encoded only 32 saliency-selected patches ($r_0$=13, $r_1$=13, $r_2$=6), corresponding to 12\% of the candidate set, yet achieved a PLCC of 0.873 on $\text{LSVQ}_{1080p}$ and 0.927 on KoNViD-1k, matching or outperforming methods while reducing computation by an order of magnitude.

\subsection{Qualitative Analysis of Learned Saliency}  
In \cref{fig:qualitative}, we visualized learned VQA-specific saliency for eight consecutive frames from three sample videos, alongside the corresponding input frames, overlay visualizations, and saliency maps from UNISAL \cite{unisal}. 

The learned saliency consistently highlighted semantically meaningful regions across time. Compared to UNISAL, which produced sharper and more concentrated responses, the VQA-specific saliency is diffused over multiple semantically relevant areas within the scene. This behavior aligns with perceptual findings that distortions in semantically meaningful regions contribute more strongly to perceived quality degradation \cite{topiq}. Additional qualitative results on more diverse content are provided in the Supplementary Material.

\subsection{Efficiency of HFVQA}
We reported the computational cost of HFVQA in terms of GFLOPs and end-to-end wall-clock latency in \cref{tab:flops_time}, alongside SOTA NR VQA models. HFVQA achieved manageable computational cost and latency that enable deployment on consumer-grade hardware, which is made possible by saliency-guided top-$k$ selection of ST patches.

Compared to recent efficient NR VQA methods such as FastVQA, FasterVQA, KVQ, and MVQA, HFVQA incurs higher computational cost for two main reasons. First, we employ a VideoPrism \cite{videoprism} encoder based on ViViT-B \cite{vivit} to utilize strong spatio-temporal priors from pretrained video foundation models. Unlike fragment-based preprocessing, HFVQA preserves the local spatio-temporal structure to remain compatible with ViFM pretraining. Second, these methods process 32 uniformly sampled frames per video (3.2 FPS for 10-second, 30 FPS clips), whereas HFVQA operates at 10 FPS, retaining denser temporal cues that directly drive its significant gains on temporally challenging datasets like LBVD and LIVE-VQC (\cref{tab:main_results,tab:ablations}). 

While employing a lighter ViFM backbone would further reduce GFLOPs and latency, to the best of our knowledge, publicly available pretrained checkpoints for substantially lighter ViFMs remain limited. Importantly, HFVQA is fully plug-and-play with respect to the ViFM encoder: as more efficient pretrained backbones become available, the computational cost of HFVQA can be directly reduced without modifying the framework itself. In addition, future work may explore teacher–student distillation \cite{distillation, tiny_clip, video_distill} to transfer ViFM knowledge into lighter student encoders for further efficiency gains. We also evaluated HFVQA with lighter pretrained video backbones (non-ViFMs). Results are provided in the Supplementary Material.


\begin{table}[t]
\centering
\footnotesize 
\setlength{\tabcolsep}{4pt} 
\caption{GFLOPs and millisecond latencies (median of 100 runs). Values are formatted as \textbf{GFLOPs / Latency}. Unless specified, latency is measured on an NVIDIA H200.}
\renewcommand{\arraystretch}{0.8} 
\begin{tabular}{l|ccc}
\toprule
\textbf{Methods} & \textbf{540p} & \textbf{720p} & \textbf{1080p} \\
\midrule
VSFA \cite{vsfa}          & 6440 / 1506 & 11426 / 2556 & 25712 / 5291 \\
PatchVQ \cite{patchvq}    & 9203 / 1792 & 13842 / 2968 & 36760 / 6556 \\
BVQA \cite{bvqa}          & 17705 / 3145 & 31533 / 7813 & 70714 / 14340 \\
MBVQA \cite{mbvqa}        & 912 / 194   & 1232 / 267   & 2150 / 891 \\
\midrule
FAST-VQA \cite{fastvqa}   & 284 / 58    & 284 / 58     & 284 / 58 \\
FasterVQA \cite{fastervqa}& 70 / 36     & 70 / 36      & 70 / 36 \\
DOVER \cite{dover}        & 282 / 59    & 282 / 59     & 282 / 59 \\
CLiF-VQA \cite{clifvqa}   & 1432 / 522  & 1432 / 522   & 1432 / 522 \\
KVQ \cite{kvq}            & 353 / 93    & 353 / 93     & 353 / 93 \\
MVQA \cite{mvqa}          & 403 / 130   & 403 / 130    & 403 / 130 \\
\midrule
\textbf{HFVQA} (H200)     & 12072 / 284 & 12072 / 284  & 12072 / 284 \\ 
\textbf{HFVQA} (RTX 4090) & 12072 / 679 & 12072 / 679 & 12072 / 679 \\ 
\bottomrule
\end{tabular}
\label{tab:flops_time}
\end{table} 
\subsection{Ablation Studies} 
In all ablations, we followed the same protocol as \cref{tab:main_results} for $\text{LSVQ}_{1080p}$, KoNViD-1k, LIVE-VQC, and LBVD.

\noindent\textbf{VQA-specific saliency.} Replacing saliency-guided top-$k$ selection and saliency-weighted aggregation (\cref{eq:within_patch_pool,eq:patch_saliency_mass}) with random sampling and uniform weighting consistently degraded performance (\cref{tab:ablations}), with PLCC/SRCC drops of 0.010/0.011 on $\text{LSVQ}_{1080p}$, 0.024/0.022 on LIVE-VQC, 0.009/0.009 on KoNViD-1k, and 0.013/0.011 on LBVD, confirming that the learned saliency selects and weights quality-relevant regions effectively.

\noindent\textbf{Multi-scale sampling.} Restricting HFVQA to scale subsets and retraining (\cref{tab:ablations}) showed that the original-resolution scale alone ($r=0$) reduces performance most on high-resolution content (0.039/0.041 on $\text{LSVQ}_{1080p}$), reflecting limited semantic and global coverage, while the coarsest scale alone ($r=2$) degrades fine-grained low-level cues. Adding the mid-scale ($r=0\,\&\,1$) consistently helped, and the full setting ($r=0,1,2$) was best across all datasets, capturing complementary high- and low-level quality cues.

\noindent\textbf{Temporal sampling rate.} Dropping the sampling rate from 10 to 3 FPS reduces the computational footprint of the framework by 72\% but reveals a trade-off. While the spatially dominated KoNViD-1k remained nearly unaffected (a 0.002/0.002 drop), performance on the temporally challenging LBVD dataset collapsed by 0.070/0.078, alongside a 0.025/0.023 degradation on LIVE-VQC. These results confirm that although internal sparse sampling offered substantial computational savings, capturing high-frequency temporal artifacts fundamentally requires both the dense sampling and the corresponding computational envelope maintained by the full HFVQA framework.
\begin{table}[t]
\centering
\footnotesize
\setlength{\tabcolsep}{2.0pt}
\renewcommand{\arraystretch}{0.75}
\caption{Ablation results on $\text{LSVQ}_{1080p}$, LIVE-VQC, KoNViD-1k, and LBVD, reported as PLCC / SRCC. A total of 32 ST patches are sampled, consistent with prior sections.} 
\begin{tabular}{l|c|c|c|c}
\toprule
\textbf{Setting} & $\textbf{LSVQ}_{1080p}$ & \textbf{LIVE-VQC} & \textbf{KoNViD-1k} & \textbf{LBVD} \\
\midrule

\multicolumn{5}{l}{\textbf{Patch Sampling}} \\
\midrule
Random         & 0.863 / 0.831  &  0.891 / 0.893  &  0.918 / 0.917  &  0.889 / 0.893\\
\textbf{Saliency}  &  \textbf{0.873} / \textbf{0.842} &  \textbf{0.915} / \textbf{0.905}  &  \textbf{0.927} / \textbf{0.926}  &  \textbf{0.902} / \textbf{0.904}  \\
\midrule

\multicolumn{5}{l}{\textbf{Scales}} \\
\midrule
$r=0$         &  0.834 / 0.801  & 0.894 / 0.886 & 0.914 / 0.912  &  0.885 / 0.887 \\
$r=0\&1$      &  0.857 / 0.829  &  0.902 / 0.895  &  0.922 / 0.922  &  0.894 / 0.895  \\
$r=2$          &  0.846 / 0.844  &  0.900 / 0.894  &  0.921 / 0.920  &  0.892 / 0.892 \\
$\textbf{\emph{r} = 0\&1\&2}$   &  \textbf{0.873} / \textbf{0.842} &  \textbf{0.915} / \textbf{0.905}  &  \textbf{0.927} / \textbf{0.926} &  \textbf{0.902} / \textbf{0.904}  \\ 
\midrule

\multicolumn{5}{l}{\textbf{Temporal Sampling Rate}} \\
\midrule
3 FPS          &  0.853 / 0.817  &  0.890 / 0.882  &  0.925 / 0.924  &  0.832 / 0.826  \\
\textbf{10 FPS}  &  \textbf{0.873} / \textbf{0.842} &  \textbf{0.915} / \textbf{0.905}  &  \textbf{0.927} / \textbf{0.926} &  \textbf{0.902} / \textbf{0.904}  \\
\bottomrule
\end{tabular}
\vspace{2pt}
\label{tab:ablations}
\end{table}

\section{Conclusion}
\label{sec:conclusion}
We presented HFVQA, a high-fidelity NR VQA framework that preserves original-resolution and dense temporal cues through multiscale ST patch sampling. By operating without altering the video data, HFVQA remains fully compatible with pretrained ViFMs, enabling effective transfer of large-scale spatio-temporal priors to VQA. To address the prohibitive cost of exhaustive encoding, we introduced a lightweight, end-to-end trained VQA-specific saliency mechanism that guides top-$k$ ST patch selection and aggregation. This design utilizes the observation that video quality perception is dominated by a small subset of spatio-temporal regions, allowing HFVQA to process as little as 12\% of the input video volume while maintaining SOTA performance. Extensive experiments demonstrate that preserving local spatio-temporal structure and dense temporal sampling yields consistent gains, particularly on high-resolution benchmarks, and enables a controlled compute–performance tradeoff. We believe HFVQA provides a principled framework for high-fidelity VQA that integrates pretrained ViFMs with task-specific saliency for compute-adaptive inference.

{
    \small
    \bibliographystyle{ieeenat_fullname}
    \bibliography{main}
}
\maketitlesupplementary

\section{Further Ablations} 
\subsection{HFVQA with Light Backbones}
\begin{table}[t]
\centering
\footnotesize
\caption{HFVQA backbone ablation on LSVQ, $\text{LSVQ}_{1080p}$, and LIVE-VQC, reported as PLCC/SRCC. All results follow the settings of Table 1.}
\label{tab:ablation_backbone}
\setlength{\tabcolsep}{2.0pt}
\renewcommand{\arraystretch}{0.8}
\begin{tabular}{l|cc|cc}
\toprule
\multirow{2}{*}{Methods} &
\multicolumn{2}{c|}{Intra-dataset} &
\multicolumn{2}{c}{Cross-dataset} \\
\cmidrule(lr){2-3}\cmidrule(lr){4-5}
& \textbf{LSVQ}$_{test}$
& \textbf{LSVQ}$_{1080p}$
& \textbf{KoNViD-1k}
& \textbf{LIVE-VQC} \\
\midrule
\textbf{Swin-S} &
0.878/0.876 &
0.822/0.787 &
0.831/0.838 &
0.822/0.827 \\
\textbf{VideoPrism-B} &
\textbf{0.906} / \textbf{0.903} &
\textbf{0.873} / \textbf{0.842} &
\textbf{0.891} / \textbf{0.893} &
\textbf{0.862} / \textbf{0.856} \\
\bottomrule
\end{tabular}

\end{table}
 
A significant portion of the computational cost of HFVQA originates from the ViFM encoder, VideoPrism-B, which is based on the ViViT-B factorized architecture. HFVQA adopts ViFMs to leverage the strong general-purpose representations obtained through large-scale video pretraining. However, currently available ViFMs are computationally heavy, which directly affects the overall cost of the framework. Importantly, HFVQA itself is not tied to a specific backbone and can be instantiated with different video encoders in a plug-and-play manner.

To examine the effect of replacing the ViFM encoder with a lighter alternative, we retrained HFVQA using a substantially lighter backbone, Swin-S \cite{videoswin}, pretrained on ImageNet-1k \cite{imagenet} and Kinetics-400 \cite{kay2017kinetics}. This variant is referred to as \textbf{HFVQA-Swin-S}. All other settings remain identical to the original configuration, including the 10 FPS temporal sampling rate.

As shown in \cref{tab:ablation_backbone}, replacing the ViFM encoder with Swin-S results in noticeable performance degradation. On $\text{LSVQ}_{test}$ and $\text{LSVQ}_{1080p}$, PLCC/SRCC decreases by \textbf{0.028/0.027} and \textbf{0.051/0.055}, respectively. The drop was also pronounced under cross-dataset evaluation, where PLCC/SRCC decreases by \textbf{0.060/0.055} on KoNViD-1k and \textbf{0.040/0.039} on LIVE-VQC.

The computational cost of HFVQA-Swin-S is substantially lower, requiring only \textbf{1593 GFLOPs}. However, the results indicate that the stronger spatio-temporal representations provided by VideoPrism-B, together with its large-scale pretraining, are critical for achieving the best performance within the HFVQA framework.

Overall, naively replacing the ViFM encoder with a lightweight backbone leads to a significant performance drop. We expect that more efficient HFVQA variants can be realized as lightweight yet capable video foundation models become available. Similar progress has been observed in the image domain, where heavy CLIP-based models \cite{clip} have been followed by lighter but competitive alternatives such as MobileCLIP family \cite{mobileclip, mobileclip2} and TinyCLIP \cite{tiny_clip}.

\section{Architecture of the Saliency Decoder} 
\begin{figure*}[t]
    \centering
    \includegraphics[width=0.99\linewidth]{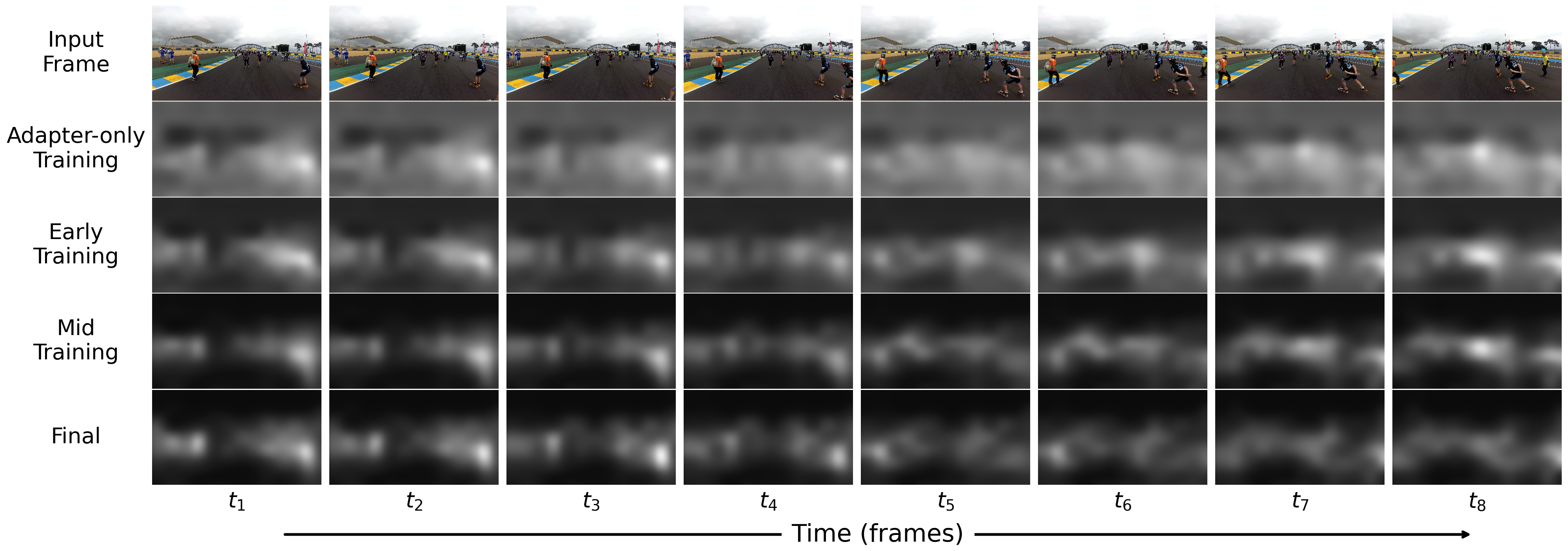}
    \caption{Consecutive video frames (top), VQA-specific saliency maps after adapter-only training (second), early training (third), mid training (fourth), and convergence (fifth). Best viewed zoomed in.    
    }
    \label{fig:saliency_progression}
\end{figure*} 
$S(\cdot)$ solves a simpler problem than full VQA, namely predicting the relative importance of dense spatio-temporal regions for VQA. Therefore, we adopt a lightweight backbone in the VQA-specific saliency module. In this lightweight regime, using a pretrained backbone is particularly important, as it provides general-purpose visual features that can be adapted to the saliency estimation task.

However, pretrained backbones typically produce abstract feature maps with reduced spatial and temporal resolution. To convert these features into dense spatio-temporal saliency maps, we introduce a lightweight adapter on top of the saliency backbone, referred to as the \textit{saliency decoder}. The role of the saliency decoder is to map the abstract feature representations into a dense probability distribution over spatio-temporal regions of the input video.

For the saliency backbone, we adopt Swin-T \cite{videoswin} pretrained on ImageNet-1k \cite{imagenet} and Kinetics-400 \cite{kay2017kinetics}. The Swin-T features are spatially downsampled by a factor of $32$ and temporally subsampled by a factor of $2$ relative to the input video. The saliency decoder progressively upsamples these features back to
the input resolution.

Specifically, the saliency decoder consists of five stages. Each stage performs bilinear spatial upsampling followed by a residual block composed of two depthwise 3D convolutions. In addition, a trilinear upsampling operation is applied at the beginning of the first stage to restore the temporal resolution. After the final stage, the resulting saliency map matches the spatial and temporal resolution of the input video. This saliency map is subsequently resized to align with the feature grids corresponding to each scale used during ST patch sampling.

Overall, the saliency decoder performs progressive upsampling and
refinement of the abstract features produced by the saliency backbone. During training, the saliency backbone remains frozen while only the saliency decoder is optimized under MOS supervision. This strategy reduces destructive interference from randomly initialized layers and stabilizes adaptation of the pretrained features \cite{lpft}.    

\section{Evolution of VQA-specific Saliency} 
We investigate whether the proposed VQA-specific saliency mechanism learns task-relevant importance patterns under MOS supervision. Training proceeds in two stages. In the first stage, only the adapter modules are optimized, including the saliency decoder and the prediction head, while the pretrained backbones remain frozen. In the second stage, all modules are jointly fine-tuned, including the saliency backbone (Swin-T), the saliency decoder, the ViFM encoder, and the prediction head.

\cref{fig:saliency_progression} visualizes the evolution of the learned saliency during training. The top row shows the input frames. The second row shows the saliency maps obtained after adapter-only training with the saliency backbone (Swin-T) frozen. The third and fourth rows show saliency maps during the early and middle stages of full fine-tuning. The final row shows the saliency maps after training convergence.

After adapter-only training, the saliency maps are highly diffuse. As training proceeds and the saliency backbone is allowed to adapt, the maps gradually become more structured and discriminative. This progression is reflected in the model performance. On the $\text{LSVQ}_{test}$, SRCC improves from \textbf{0.852} after adapter-only training to \textbf{0.891} and \textbf{0.900} during early and mid fine-tuning, respectively, reaching a final value of \textbf{0.903} after convergence.

Importantly, the saliency maps change little after mid training and converge to a stable behavior rather than fluctuating erratically across epochs. This suggests that the final saliency patterns are not incidental artifacts of training but reflect task-driven importance patterns learned from MOS supervision. As illustrated in Fig. 3 of the main paper and further examples provided in \cref{sec:additional_qualitative}, the learned saliency shares certain characteristics with conventional visual saliency while also exhibiting clear differences, which we attribute to the task-specific characteristics induced by VQA supervision.

\section{Additional Qualitative Results} 
\label{sec:additional_qualitative}
\begin{figure*}[tbp]
    \centering
    \includegraphics[width=0.86\linewidth]{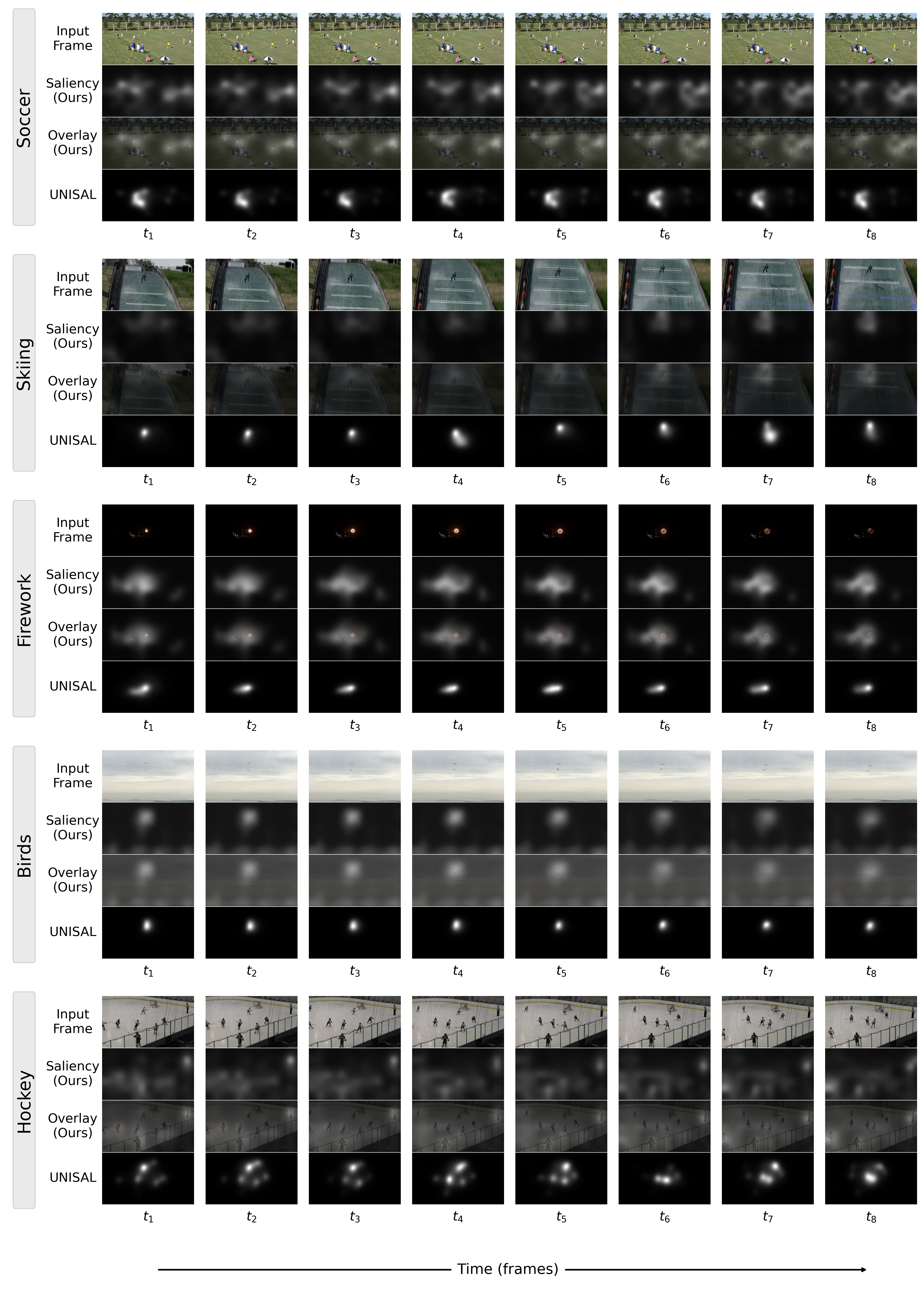}
    \caption{ Consecutive video frames (top), learned VQA-specific saliency maps (second), overlays (third), and comparison with UNISAL (bottom). Best viewed zoomed in.}
    \label{fig:qualitative_supp1}
\end{figure*} 
\begin{figure*}[tbp]
    \centering
    \includegraphics[width=0.86\linewidth]{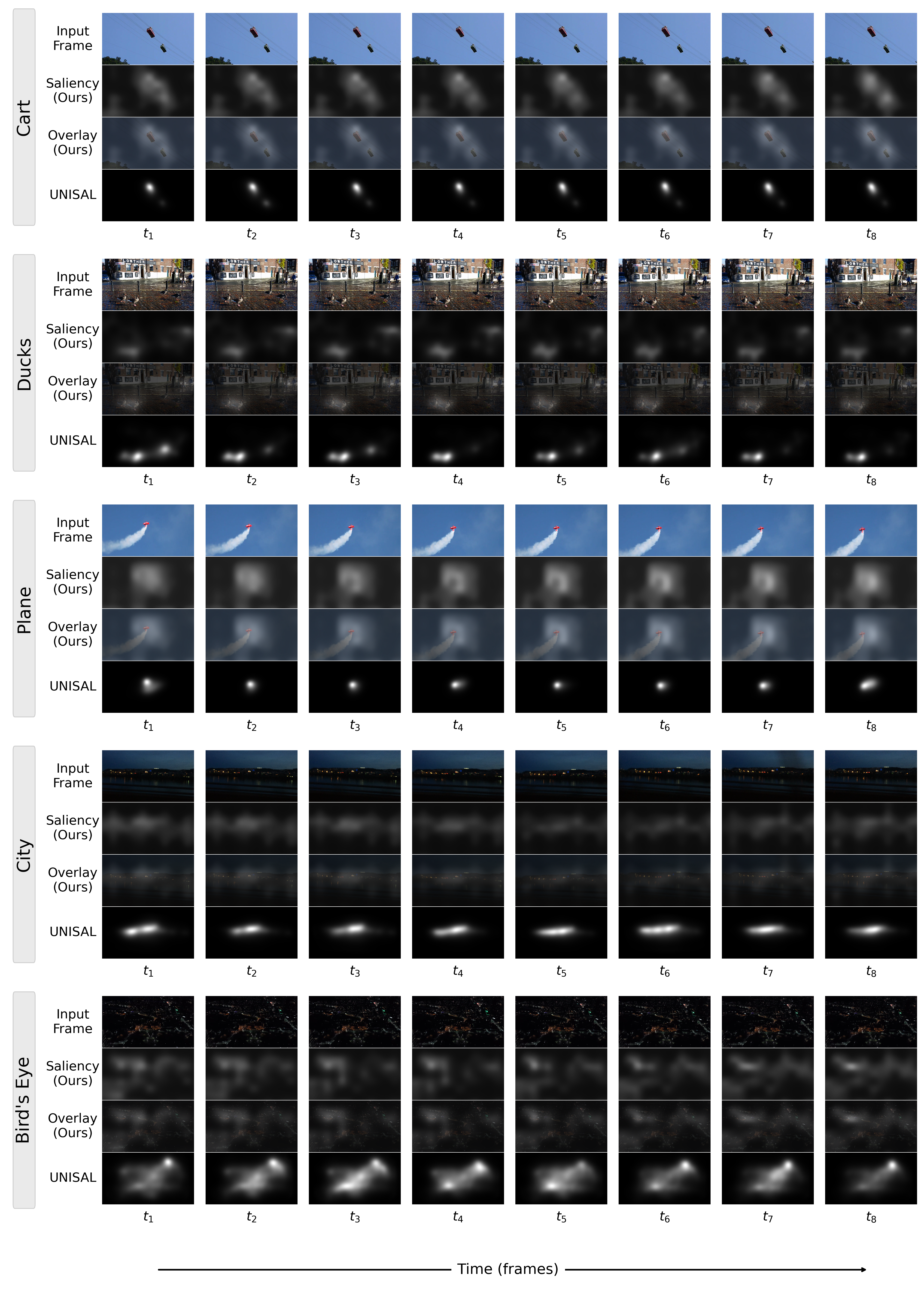}
    \caption{Consecutive video frames (top), learned VQA-specific saliency maps (second), overlays (third), and comparison with UNISAL (bottom). Best viewed zoomed in.}
    \label{fig:qualitative_supp2}
\end{figure*} 

We provided additional qualitative visualizations of the learned VQA-specific saliency in \cref{fig:qualitative_supp1,fig:qualitative_supp2}. Across diverse scenes, the learned saliency frequently emphasized semantically meaningful regions, including humans, animals, and prominent objects that are likely to influence perceived video quality.

Across consecutive frames, the saliency maps also appeared relatively consistent over time and often remain centered on dynamic scene elements. While such behavior is difficult to be fully assessed from still visualizations, this tendency suggests a possible bias toward temporally salient regions, which aligns with perceptual observations that human attention is often drawn to motion and animated content.

In scenarios where clear semantic structures were less prominent, such as low-light or low-quality scenes, the learned saliency became more spatially diffused and often concentrated around bright or visually distinct regions. This behavior suggests that when semantic cues are weak, the model may rely more heavily on basic visual structures that can reveal compression artifacts, noise, or other distortions.

Overall, these qualitative observations support the interpretation that the learned saliency captures task-relevant importance patterns rather than generic visual saliency, adapting its spatial focus depending on both scene semantics and local visual characteristics.

\end{document}